\def\year{2022}\relax
\documentclass[letterpaper]{article} 
\usepackage{aaai22}  
\usepackage{times}  
\usepackage{helvet}  
\usepackage{courier}  
\usepackage[hyphens]{url}  
\usepackage{graphicx} 
\usepackage{natbib}  
\usepackage{caption} 
\DeclareCaptionStyle{ruled}{labelfont=normalfont,labelsep=colon,strut=off} 
\usepackage{algorithm}
\usepackage{algorithmic}
\usepackage{array}
\newcommand{\PreserveBackslash}[1]{\let\temp=\\#1\let\\=\temp}
\newcolumntype{C}[1]{>{\PreserveBackslash\centering}p{#1}}
\newcolumntype{R}[1]{>{\PreserveBackslash\raggedleft}p{#1}}
\newcolumntype{L}[1]{>{\PreserveBackslash\raggedright}p{#1}}
\usepackage{multirow}

\usepackage{amsfonts,amssymb}
\usepackage{color}
\definecolor{purplefish}{RGB}{138,43,226}
\definecolor{orangefish}{RGB}{210,105,30}
\definecolor{crimson}{RGB}{220,20,60}
\usepackage{mathrsfs}
\usepackage{url}
\usepackage{subfigure}
\usepackage{graphicx}

\usepackage[colorlinks,
            linkcolor=black,
            anchorcolor=black,
            citecolor=black,
            urlcolor=black]{hyperref}
\usepackage[switch]{lineno}
\usepackage{appendix}
            
\usepackage{newfloat}
\usepackage{listings}
\floatstyle{ruled}
\newfloat{listing}{tb}{lst}{}
\floatname{listing}{Listing}
\title{XLM-K: Improving Cross-Lingual Language Model Pre-training with Multilingual Knowledge}

\author {
Xiaoze Jiang\textsuperscript{\rm 1}\footnote{Contribution during internship at Microsoft Research Asia.}, Yaobo Liang\textsuperscript{\rm 2}, Weizhu Chen\textsuperscript{\rm 3}, Nan Duan\textsuperscript{\rm 2}\\

}
\affiliations{
    \textsuperscript{\rm 1}Beihang University, Beijing, China\\
    \textsuperscript{\rm 2}Microsoft Research Asia, Beijing, China\\ 
    \textsuperscript{\rm 3}Microsoft Azure AI, Redmond, WA, USA \\
    xzjiang@buaa.edu.cn, \{yalia, wzchen, nanduan\}@microsoft.com
}

\usepackage{bibentry}

\begin{document}

\maketitle

\begin{abstract}
Cross-lingual pre-training has achieved great successes using monolingual and bilingual plain text corpora. However, most pre-trained models neglect multilingual knowledge, which is language agnostic but comprises abundant cross-lingual structure alignment. In this paper, we propose XLM-K, a cross-lingual language model incorporating multilingual knowledge in pre-training. XLM-K augments existing multilingual pre-training with two knowledge tasks, namely Masked Entity Prediction Task and Object Entailment Task. We evaluate XLM-K on MLQA, NER and XNLI. Experimental results clearly demonstrate significant improvements over existing multilingual language models. The results on MLQA and NER  exhibit the superiority of XLM-K in knowledge related tasks. The success in XNLI shows a better cross-lingual transferability obtained in XLM-K. What is more, we provide a detailed probing analysis to confirm the desired knowledge captured in our pre-training regimen. The code is available at \url{https://github.com/microsoft/Unicoder/tree/master/pretraining/xlmk}.
\end{abstract}

\section{Introduction}

Recent development of pre-trained language model \cite{devlin2019bert,liu2019roberta} has inspired a new surge of interest in the cross-lingual scenario, such as Multilingual BERT \cite{devlin2019bert} and XLM-R \cite{conneau2020unsupervised}. 
Existing models are usually optimized for masked language modeling (MLM) tasks \cite{devlin2019bert} and translation tasks \cite{conneau2019cross} using multilingual data. 
However, they neglect the knowledge across languages, such as entity resolution and relation reasoning.  In fact, the knowledge conveys similar semantic concepts and similar meanings across languages \cite{vulic-moens-2013-cross,chen2020cross}, which is essential to achieve cross-lingual transferability.
Therefore, how to equip pre-trained models with knowledge has become an underexplored but critical challenge for multilingual language models.


Contextual linguistic representations in language models are ordinarily trained using unlabeled and unstructured corpus, without the consideration of explicit grounding to knowledge \cite{fevry2020entities,xiong2020pretrained,fan2021discovering}, such as entity and relation. On one side, the structural knowledge data is abundant and could be a great complement to the unstructured corpus for building a better language model. Many works have demonstrated its importance via incorporating basic knowledge into monolingual pre-trained models \cite{zhang2019ernie,staliunaite2020compositional,zhang2020grounded, wang2020kepler}. On another side, knowledge is often language agnostic, e.g. different languages share the same entity via different surface forms. This can introduce a huge amount of alignment data to learn a better cross-lingual representation \cite{cao-etal-2018-joint}.
However, there are few existing works on exploring the multilingual entity linking and relation in the cross-lingual setting for pre-training \cite{huang2019unicoder,yang2020alternating}. For example, the de-facto cross-lingual pre-training standard, i.e. MLM \cite{devlin2019bert}  plus TLM \cite{conneau2019cross}, learns the correspondences between the words or sentences across the languages,  neglecting the diverse background cross-lingual information behind each entity.

To address this limitation, we propose XLM-K, a cross-lingual language model incorporating multilingual knowledge in pre-training. The knowledge is injected into the XLM-K via two additional pre-trained tasks, i.e. \textit{masked entity prediction} task and \textit{object entailment} task. These two tasks are designed to capture the knowledge from two aspects: description semantics and structured semantics. 
Description semantics encourage the contextualized entity embedding in a sequence to be linked to the long entity description in the multilingual knowledge base (KB).
Structured semantics, based on the triplet knowledge $<$\textit{subject, relation, object}$>$, connect cross-lingual subject and object based on their relation and descriptions, in which  the \textit{object} is entailed by the joint of the \textit{subject} and the \textit{relation}. The \textit{object} and \textit{subject} are both represented by their description from the KB. To facilitate the cross-lingual transfer ability, on one hand, the entity and its description are from different languages. On the other hand, the textual contents of the subject and object also come from a distinct language source. We employ the contrastive learning \cite{he2020momentum} during pre-training to make XLM-K distinguish a positive knowledge example from a list of negative knowledge examples. 




There are three main contributions in our work: 

$\bullet$ \ \  As the first attempt, we achieve the combination between the textual information and knowledge base in cross-lingual pre-training by proposing two knowledge related and cross-lingual pre-training tasks. The knowledge, connected via different languages, introduces additional information for learning a better multilingual representation.

$\bullet$ \ \   We evaluate XLM-K on the entity-knowledge related downstream tasks, i.e. MLQA and NER, as well as the standard multilingual benchmark XNLI. Experimental results show that XLM-K achieves new state-of-the-art results in the setting without bilingual data resource. The improvements in MLQA and NER show its superiority on knowledge related scenarios. The results on XNLI demonstrate the better cross-lingual transferability in XLM-K.

$\bullet$ We further perform a probing analysis \cite{petroni2019language} on XLM-K, clearly reflecting the desired knowledge in the pre-trained models.

\section{Related Work}

\paragraph{Cross-Lingual Pre-training} Works on cross-lingual pre-training have achieved a great success in multilingual tasks. Multilingual BERT \cite{devlin2019bert} trains a BERT model based on multilingual masked language modeling task on the monolingual corpus. XLM-R \cite{conneau2020unsupervised} further extends the methods on a large scale corpus. These models only use monolingual data from different languages. To achieve cross-lingual token alignment, XLM \cite{conneau2019cross} proposes translation language modeling task on parallel corpora. Unicoder \cite{huang2019unicoder} presents several pre-training tasks upon parallel corpora and InfoXLM \cite{chi2020infoxlm} encourages bilingual sentence pair to be encoded more similar than the negative examples, while ERNIE-M \cite{ouyang2020ernie} learns semantic alignment among multiple languages on monolingual corpora. These models leverage bilingual data to achieve better cross-lingual capability between different languages. Our method explores cross-lingual knowledge base as a new cross-lingual supervision.



\paragraph{Knowledge-Aware Pre-training} Recent monolingual works, incorporating basic knowledge into  monolingual pre-trained models, result to a better performance in downstream tasks \cite{rosset2020knowledge}. For example, 
some works introduce entity information via adding a knowledge specific model structure \cite{broscheit2019investigating,zhang2019ernie,fevry2020entities}.  Others consider the relation information captured in the knowledge graph triples \cite{hayashi2020latent,zhang2020grounded,wang2020k,liu2020k}. Meanwhile,  \citet{xiong2020pretrained,fevry2020entities}  equip monolingual language model with diverse knowledge without extra parameters. These works are almost in monolingual domain without the  consideration of cross-lingual information, while the cross-lingual knowledge is learned by our model via the proposed tasks. Moreover, the standard operation of aforementioned works are mostly based on the entity names. The entity names are masked and then predicted by the model, namely the MLM task is conducted on the masked entity names. While we predict the entity description for the purpose of disambiguation of different entities with the same entity name (detailed in Sec. \ref{sec:mep}). It can help our model learn more fine-grained knowledge.

\paragraph{Word Embedding and KB Joint Learning} Many works leverage word embedding from text corpus to generate better KB embedding. \citealt{wang2014knowledge,yamada2016joint,cao2017bridge} utilize the texts with entity mention and entity name to align word embedding and entity embedding. \citealt{toutanova2015representing,han2016joint,wu2016knowledge,wang2016text} utilize the sentences with two entity mentions as relation representation to generate better entity and relation embedding. These works mainly target to generate better graph embedding with English corpus and each entity will have a trainable embedding. Our methods focus on training a better contextualized representation for multiple languages. Meanwhile, the entity representation is generated by Transformer \cite{vaswani2017attention} model, which could further align the textual and KB embedding, as well as achieving the less trainable parameters. For cross-lingual word embeddings, most of works rely on aligned words or sentences \cite{ruder2019survey}. \citealt{cao2018joint,pan-etal-2019-cross,chen2020cross} replace the entity mention to a special entity tag and regularize one entity's different mentions in different languages to have similar embedding. \citealt{vulic-moens-2013-cross} use topic tag of Wikipedia pages to improve the cross-lingual ability. We also utilize entity mention in different languages as cross-lingual alignment supervision. Different from these works, we further exploit relation information to enhance the entity representation. What's more, we generate entity representation by Transformer model instead of training the separate embedding for special entity tag.

\begin{figure*}[t]
    \centering
    \subfigure[Masked Entity Prediction Task]{
    
        \begin{minipage}[b]{\textwidth}
            \includegraphics[width=1\textwidth]{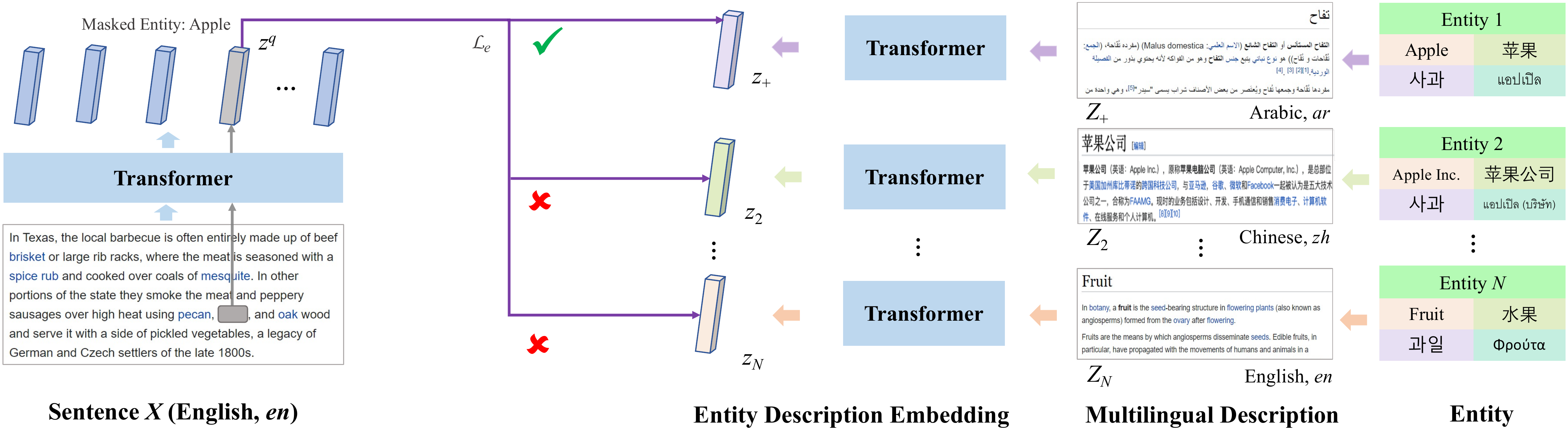}
        \end{minipage}
    }
    \subfigure[Object Entailment Task]{
    
        \begin{minipage}[b]{\textwidth}
            \includegraphics[width=1\textwidth]{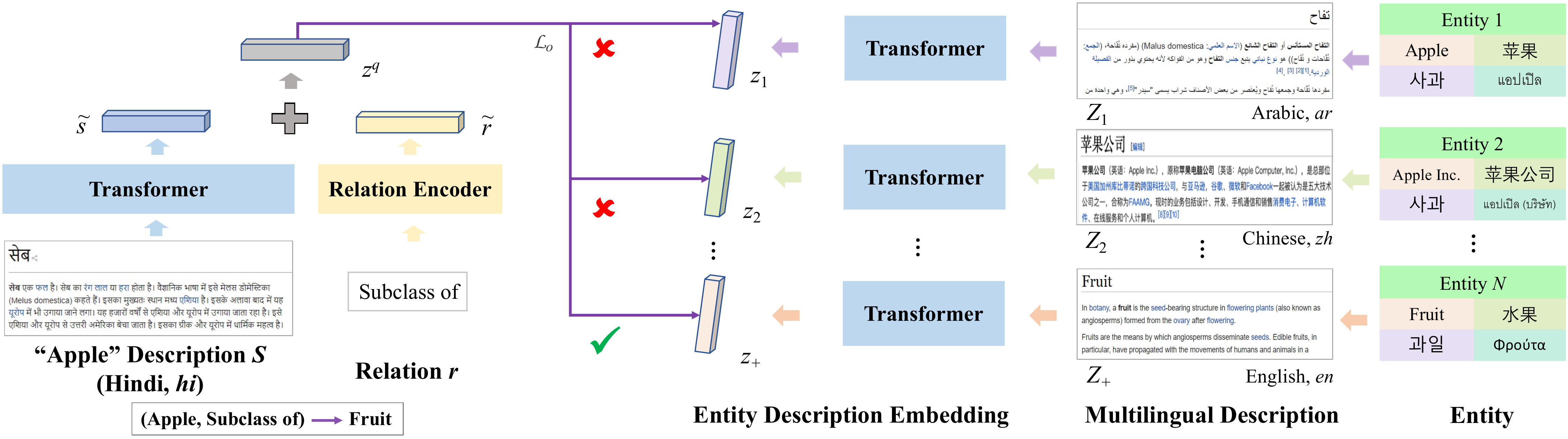}
        \end{minipage}
    }
    \caption{XLM-K mainly consists of two cross-lingual pre-training tasks: (a) Masked Entity Prediction recognizes the masked entity with its knowledge description (the entity {\it Apple} is masked in sentence $X$); (b) Object Entailment predicts the textual contents of {\it object} with the combination of {\it subject} and {\it relation}. All the Transformers are with shared parameters.} 
    \label{fig:model}
\end{figure*}

\section{Methodology}

We first present the knowledge construction strategy. Then, we introduce our two knowledge-based pre-training tasks and the training objective.


\subsection{Knowledge Construction}
\label{know-cons}


We use \textit{Wikipedia} and \textit{Wikidata} \cite{vrandevcic2014wikidata} as the data source. More details refer to Appendix.


\noindent \textbf{Knowledge Graph} A knowledge graph is a set of triplets in form $<$\textit{subject, relation, object}$>$. We use Wikidata as our knowledge base. The triplets of Wikidata are extracted from Wikipedia and each Wikipedia page corresponding to an entity in WikiData. WikiData contains 85 million entities and 1304 relations. They formed 280 million triplets.

\noindent \textbf{Entity Mention} For a sentence with $l$ words, $X=(x_1, x_2, ..., x_l)$, a mention $(s, t, e)$ means the sub-sequence $(x_s, x_{(s+1)}, ..., x_t)$ corresponding to entity $e$. In our work, we use Wikipedia as data source.  For each anchor in Wikipedia, it provides the link to the Wikipedia page of this entity, which can be further mapped to a unique entity in Wikidata. Wikipedia pages are from 298 languages, and every around 64 tokens contains an anchor. 









\noindent \textbf{Multilingual Entity Description} We treat a Wikipedia page as the description of its corresponding entity in WikiData. Since Wikipedia contains multiple languages, an entity may have multiple descriptions and they come from different languages. For each page, we only keep its first 256 tokens as its description. As shown in Figure \ref{fig:model}, $N$ multilingual entity descriptions form the candidate list $\mathcal{Z}=\{Z_{1},Z_{2},...,Z_{N}\}$.



\subsection{Masked Entity Prediction Task}
\label{sec:mep}


Masked entity prediction task is to encourage the contextualized entity embedding in a sequence to predict the long entity description in the multilingual knowledge base (KB), rather than the prediction of entity name. It can help the disambiguation of the different entity with the same entity name. For example, as shown in Figure \ref{fig:model}.a, the entity name of {\it Apple} and {\it Apple Inc.} are the same in Korean. It helps XLM-K learn the diverse implicit knowledge behind the mentioned words.





Given a sentence $X=(x_1,x_2,...,x_l)$ from the cross-lingual corpus, where $X$ is a sentence with $l$ words from language $u^{lg}$ (e.g. $u^{lg}$ is {\it en}, shown in Figure  \ref{fig:model}.a), and a masked mention $(s, t, e)$ (replaced by {\ttfamily [MASK]}), the task is to recognize the positive example $Z_{+}$ from a candidate list $\mathcal{Z}$, which contains distracting pages from multiple languages but associated with other entities. $Z_{+}=(z_1,z_2,...,z_m)$ is the description of entity $e$ with $m$ words from language $t^{lg}$ (e.g. $t^{lg}$ is {\it ar}, shown in  Figure \ref{fig:model}.a). Note that the description $Z_{+}$ (with a maximum 256 tokens) is extracted from the related Wikipedia page of entity $e$. After $X$ being fed into the Transformer encoder, the final hidden state of $x_s$, denotes $x_s^t$, and the {\ttfamily [CLS]} from $Z_{+}$, denotes $\widetilde{z}$, are further fed into a non-linear projection layer \cite{chen2020simple}, respectively:
\begin{equation}
z^{q} = W_2 ReLU(W_1 x_s^t)
\end{equation}
\begin{equation}
z_{+} = W_4 ReLU(W_3 \widetilde{z})
\end{equation}
where $W_1, W_3 \in \mathbb{R}^{d_w \times d_p}$ and $W_2, W_4 \in \mathbb{R}^{d_p \times d_w}$. Then the masked entity prediction loss $\mathcal{L}_e$ can be calculated  by Eq. \ref{eq:contrastive}.


\subsection{Object Entailment Task}

The masked entity prediction task enriches XLM-K with sentence-level semantic knowledge, while object entailment task is designed to enhance the structured relation knowledge. As shown in Figure \ref{fig:model}.b, given the {\it subject} and {\it relation}, the model is forced to select the {\it object} from the candidate list. For the purpose of entity disambiguation (explained in Sec. \ref{sec:mep}), the representations of {\it subject} and {\it object} are also from the long entity description.



Formally, given the subject entity's description sentence $S=(s_1, s_2, ..., s_l)$ with $l$ words from language $u^{lg}$ (e.g. $u^{lg}$ is {\it hi}, shown in Figure \ref{fig:model}.b), the object entity's description sentence $Z_{+}=(z_1,z_2,...,z_m)$ with $m$ words from language $t^{lg}$ (e.g. $t^{lg}$ is {\it en}, shown in Figure \ref{fig:model}.b) and their relation $r$ (language agnostic), the task is to predict the object $Z_{+}$ from a cross-lingual candidate list $\mathcal{Z}$, based on $S$ and $r$. Firstly, the relation $r$ is fed into the \textit{Relation Encoder} (a look-up layer to output the relation embedding), and subject entity description sentence $S$ and object entity description sentence $Z_{+}$ is fed into a separate Transformer encoder. We can get the encoded relation $\widetilde{r}$, the whole representation of subject entity description sentence $\widetilde{s}$ and object entity description sentence $\widetilde{z}$, based on their {\ttfamily [CLS]} in the last layer. The joint embedding of $\widetilde{s}$ and $\widetilde{r}$ is constructed as follows:
\begin{equation}
z^q = W_6 ReLU(W_5 (\widetilde{s} + \widetilde{r}))
\end{equation}
where $W_5 \in \mathbb{R}^{d_w \times d_p}$ and $W_6 \in \mathbb{R}^{d_p \times d_w}$ are trainable weights. The object $\widetilde{z}$ is also encoded by a non-linear projection layer:
\begin{equation}
z_{+} = W_8 ReLU(W_7 \widetilde{z})
\end{equation}
where $W_7 \in \mathbb{R}^{d_w \times d_p}$ and $W_8 \in \mathbb{R}^{d_p \times d_w}$. The object entailment loss $\mathcal{L}_o$  is calculated  by Eq. \ref{eq:contrastive}.


\subsection{Joint Pre-training Objective}
\label{sec:loss}


Although we can have different loss functions to optimise XLM-K,  we choose {\it contrastive learning} due to its promising results in both visual representations \cite{he2020momentum,chen2020simple} and cross-lingual pre-training \cite{chi2020infoxlm,pan2020multilingual}. Intuitively, by distinguishing the positive sample from the negative samples using the contrastive loss, the model stores expressive knowledge acquired from the structure data. Formally, the loss can be calculated as:
\begin{equation}
\mathcal{L}_{e} ({\rm and} \  \mathcal{L}_{o})= - {\rm log} \frac{exp(z^q z_{+})}{\sum_{k=1}^{N} exp(z^q z_{k})}
\label{eq:contrastive}
\end{equation}
where $z_{+}$ is the positive sample, $z_k$ is the $k$-th  candidate sample (encoded by the same way like $z_{+}$) and $N$ is the size of the candidate list $\mathcal{Z}$.
To avoid catastrophic forgetting of the learned knowledge from the previous training stage, we preserve the multilingual masked language modeling objective (MLM) \cite{devlin2019bert}, denotes $\mathcal{L}_{\rm MLM}$. As a result, the optimization objective of XLM-K is defined as:
\begin{equation}
\mathcal{L} =  \mathcal{L}_{\rm MLM} + \mathcal{L}_e  + \mathcal{L}_o
\label{eq:loss}
\end{equation}







\section{Experiments}

 
In this section, we will introduce implementation details of XLM-K, then, evaluate the performance of XLM-K on the downstream tasks. Lastly, we conduct probing experiments on the pre-trained models to verify the knowledge can be stored via the proposed tasks.

\subsection{Implementation Details}

\textbf{Data and Model Structure} For the multilingual masked language modeling task, we use Common Crawl dataset \cite{wenzek2019ccnet}. The  Common Crawl dataset is crawled from the whole web without restriction, which contains all the corpus from the Wikipedia. For the proposed two tasks, we use the corpus for the top 100 languages with the largest Wikipedias. The settings to balance the instances from different languages are the same as XLM-R$_{base}$ \cite{conneau2020unsupervised}. The architecture of XLM-K is set as follows: 768 hidden units, 12 heads, 12 layers, GELU activation, a dropout rate of 0.1, with a maximal input length of 256 for the proposed knowledge tasks, and 512 for MLM task.


\noindent \textbf{Details of Pre-training} We initialize the model with XLM-R$_{base}$ \cite{conneau2020unsupervised} (was trained on Common Crawl), and conduct continual pre-training with the gradient accumulation of 8,192 batch size. We utilize Adam \cite{kingma2015adam} as our optimizer. The learning rate starts with 10k warm-up steps and the peak learning rate is set to 3e-5. The size of candidate list size $N=32$k. The candidate list is implemented as a queue, randomly initialized at the beginning of the training stage and updated by the newly encoded entities (more details refer to Appendix). The pre-training experiments are conducted using 16 V100 GPUs.


\noindent \textbf{Details of Fine-Tuning} We follow \citealt{liang2020xglue} in these fine-tuning settings. In detail, we use Adam optimizer with warm-up and only fine-tune XLM-K on the English training set. \textbf{F}or MLQA, we fine-tune 2 epochs, with the learning rate set as 3e-5 and batch size of 12. \textbf{F}or NER, we fine-tune 20 epochs, with the learning rate set as 5e-6 and batch size of 32. \textbf{F}or XNLI, we fine-tune 10 epochs and the other settings are the same as for NER.  \textbf{W}e test all the fine-tuned models on dev split of all languages for each fine-tuning epoch and select the model based on the best average performance on the dev split of all languages. To achieve a convincing comparison, we run the fine-tuning experiments with 4 random seeds and report both the average and maximum results on all downstream tasks. We also run our baseline XLM-R$_{base}$ with the same 4 seeds and report average results.

\noindent \textbf{Details of Probing} Following \citet{petroni2019language}, we conduct probing analysis directly on the pre-trained models without any fine-tuning. The probing corpus are from four sources: Google-RE\footnote{\url{https://code.google.com/archive/p/relation-extraction-corpus/}}, T-REx \cite{elsahar2018t}, ConceptNet \cite{speer2012representing} and SQuAD \cite{rajpurkar2016squad}. Except that ConceptNet tests for commonsense knowledge, others are all designed to probe Wiki-related knowledge.
\begin{table*}[t] 
\centering
\resizebox{1\textwidth}{!}{
\begin{tabular}{lcccccccc}
\hline                       
Model & en & es & de & ar & hi & vi & zh & Avg \\ 
\hline  
mBERT \cite{lewis020mlqa} & 77.7/65.2 & 64.3/46.6 & 57.9/44.3 & 45.7/29.8 & 43.8/29.7 & 57.1/38.6 & 57.5/37.3 & 57.7/41.6 \\
XLM \cite{lewis020mlqa} & 74.9/62.4 & 68.0/49.8 & 62.2/47.6 & 54.8/36.3 & 48.8/27.3 & 61.4/41.8 & 61.1/39.6 & 61.6/43.5 \\
mBERT + Post-Pretraining Alignment \cite{pan2020multilingual} & 79.8/ \ \ - \ \ \   & 67.7/  \ \ - \ \ \   & 62.3/  \ \ - \ \ \   & 53.8/  \ \ - \ \ \   & 57.9/  \ \ - \ \ \   &  \ \ - \ \ \   /\ \ \ - \ \ \   & 61.5/  \ \ - \ \ \   & 63.8/  \ \ - \ \ \   \\
Unicoder \cite{huang2019unicoder} & 80.6/\ \ \ - \ \ \   & 68.6/\ \ \ - \ \ \  & 62.7/\ \ \ - \ \ \  & 57.8/\ \ \ - \ \ \  & 62.7/\ \ \ - \ \ \  & 67.5/\ \ \ - \ \ \  & 62.1/\ \ \ - \ \ \ & 66.0/\ \ \ - \ \ \   \\
XLM-R$_{base}$ \cite{conneau2020unsupervised} & 80.1/67.0 & 67.9/49.9 & 62.1/47.7 & 56.4/37.2 & 60.5/44.0 & 67.1/46.3 & 61.4/38.5 & 65.1/47.2 \\

\hline  
MEP ({\it avg}) & 80.6/67.5  & 68.7/50.9 & 62.8/48.2 & 59.0/39.9 & 63.1/46.1 & 68.2/47.5 & 62.1/38.1 & 66.4/48.3 \\
OE ({\it avg}) & 80.8/67.8 & 69.1/51.2 & 63.2/48.6 & 59.0/39.6 & 63.7/46.3 & 68.5/47.3 & 63.0/39.5 & 66.7/48.6 \\
\hline 
\textbf{XLM-K} ({\it avg}) & 80.8/67.7 & 69.3/51.6 & 63.2/48.9 & 59.8/40.5 & 64.3/46.9 & 69.0/48.0 & 63.1/38.8 & 67.1/48.9 \\
\textbf{XLM-K} ({\it max}) &  80.8/67.9 & 69.2/52.1 & 63.8/49.2 & 60.0/41.1 & 65.3/47.6 & 70.1/48.6 & 63.8/39.7 & \textbf{67.7}/\textbf{49.5} \\
\hline  
\hline
{\it with Bilingual Data} \\
\hline 
InfoXLM \cite{chi2020infoxlm}& 81.3/68.2 & 69.9/51.9 & 64.2/49.6 & 60.1/40.9 & 65.0/47.5 & 70.0/48.6 & 64.7/41.2 & 67.9/49.7 \\
ERNIE-M \cite{ouyang2020ernie} & 81.6/68.5 & 70.9/52.6 & 65.8/50.7 & 61.8/41.9 & 65.4/47.5 & 70.0/49.2 & 65.6/41.0 & \textbf{68.7}/\textbf{50.2} \\
\hline
\end{tabular}}
\caption{The results of MLQA F1/EM (exact match) scores on each language. The models in the second block are our ablation models MEP and OE. We run our model and ablation models four times with different seeds, where {\it avg} means the average results and {\it max} means the maximum results selected by the Avg metrics. Statistical significance test (Standard Deviation $\sigma$, {\it value}$\pm\sigma$) for XLM-K({\it avg}) on Avg: 67.1$\pm$0.3 / 48.9$\pm$0.2.}
\label{table:mlqa}  
\end{table*} 
\begin{table}[t] 
\centering
\resizebox{0.48\textwidth}{!}{
\begin{tabular}{lccccc}
\hline                       
Model & en & es & de & nl & Avg \\ 
\hline  
mBERT from \cite{liang2020xglue} & 90.6 & 75.4 & 69.2 & 77.9 & 78.2 \\
XLM-R$_{base}$ from \cite{liang2020xglue} &  90.9 & 75.2 & 70.4 & 79.5  & 79.0\\
\hline 
MEP ({\it avg}) & 90.6 & 75.6 & 72.3 & 80.2 & 79.6 \\
OE ({\it avg}) & 90.9 & 76.0 & 72.7 & 80.1 & 79.9 \\
\hline 
\textbf{XLM-K} ({\it avg}) & 90.7 & 75.2 & 72.9 & 80.3 &  79.8  \\
\textbf{XLM-K} ({\it max}) & 90.7 &  76.6 & 73.3 & 80.0 & \textbf{80.1}\\
\hline  
\end{tabular}}  
\caption{The results of NER F1 scores on each language. The models in the second block are our ablation models. The meaning of {\it avg} and {\it max} are the same as Table  \ref{table:mlqa}. Statistical significance test (Standard Deviation $\sigma$, {\it value}$\pm\sigma$) for XLM-K({\it avg}) on Avg: 79.8$\pm$0.2.}
\label{table:ner}  
\end{table}

\subsection{Downstream Task Evaluation}
\label{downstream}

To evaluate the performance of our model using downstream tasks, we conduct experiments on MLQA, NER and XNLI. MLQA and NER are entity-related tasks, and XNLI is a widely-used cross-lingual benchmark. Without using bilingual data in pre-training, we achieve new state-of-the-art results on these three tasks. For the convenience of reference, we display the results of the bilingual data relevant methods, namely the recently released models InfoXLM \cite{chi2020infoxlm} and ERNIE-M \cite{ouyang2020ernie}, in  Table \ref{table:mlqa} and Table \ref{table:xnli} and omit the analysis. Applying bilingual data resources to XLM-K is left as future work. In the following section, {\bf MEP} means the ablation model of Masked Entity Prediction + MLM and {\bf OE} means Object Entailment + MLM.

\noindent \textbf{MLQA} MLQA \cite{lewis020mlqa} is a multilingual question answering dataset, which covers 7 languages including \textit{English}, \textit{Spanish}, \textit{German}, \textit{Arabic}, \textit{Hindi}, \textit{Vietnamese} and \textit{Chinese}. As a big portion of questions in MLQA are factual ones, we use it to evaluate XLM-K that is pre-trained using the multilingual knowledge. 

The results on MLQA are shown in Table \ref{table:mlqa}. Since F1 and EM scores have similar observations, we take F1 scores for analysis:

(1) \textit{The Effectiveness of XLM-K.} For the \textit{avg} report, XLM-K achieves 67.1 averaged accuracy on F1 score, outperforming the baseline model XLM-R$_{base}$ by 2.0. For the \textit{max} report, the model can further obtain 0.6 additional gain over the \textit{avg} report. This clearly illustrate the superiority of XLM-K on MLQA. In addition, the model MEP and OE provide 1.3 and 1.6 improvements over XLM-R$_{base}$, respectively, which reveals that each task can capture MLQA's task-specific knowledge successfully.

(2) \textit{The Ablation Analysis of XLM-K.} The models in the second block are the ablation models. Compared with the ablation models, XLM-K outperforms each model by 0.7 and 0.4 on Avg metrics. It indicates that the masked entity prediction and object entailment has complementary advantages on MLQA task, and the best result is achieved when using them together.



\noindent \textbf{NER} The cross-lingual NER \cite{liang2020xglue} dataset covers 4 languages, including \textit{English}, \textit{Spanish}, \textit{German} and \textit{Dutch}, and 4 types of named entities, namely \textit{Person}, \textit{Location}, \textit{Organization} and \textit{Miscellaneous}. 
As shown in Table \ref{table:ner}, 
compared with baseline model XLM-R$_{base}$, XLM-K improves the Avg score to 79.8 on average and 80.1 on maximum. It verifies the effectiveness of XLM-K when solving NER task. Meanwhile, the results of MEP and OE are also increased by 0.6 and 0.9 on Avg F1 score. It displays that the entity-related pre-training task has significant improvements on the entity-related downstream tasks.






\noindent \textbf{XNLI} The XNLI \cite{conneau2018xnli} is a popular evaluation dataset for cross-lingual NLI which contains 15 languages. It's a textual inference tasks and not rerely relied on knowledge base. 
We present the results in Table \ref{table:xnli} with following observations:

(1) \textit{The Effectiveness of XLM-K.} Although XNLI is not an entity or relation -aware multilingual task, our model obtains a 0.6 gain comparing to the baseline model XLM-R$_{base}$. Each ablation model of MEP and OE improve by 0.4. These gains are marginal compared to MLQA and NER. This shows that our model is mainly works on knowledge-aware tasks. On other tasks, it won't harm the performance and even could marginally help.





(2) \textit{The Ablation Analysis of XLM-K.} The ablation models of XLM-K on XNLI have similar results on XNLI, which increasing by 0.4 compared to the XLM-R$_{base}$ baseline 74.2. It proves each task has its own contribution to the overall improvements. Meanwhile, the ablation models still have 0.2 gap to the XLM-K, which implies the advantages towards the combination of these two tasks.

\begin{table*}[t] 
\centering
\resizebox{\textwidth}{!}{
\begin{tabular}{lcccccccccccccccc}
\hline                       
Model & en &  fr &  es &  de &  el &  bg &  ru &  tr &  ar &  vi &  th &  zh &  hi &  sw &  ur & Avg \\ 
\hline  
mBERT \cite{conneau2020unsupervised}  & 82.1 & 73.8 & 74.3 & 71.1 & 66.4 & 68.9 & 69.0 & 61.6 & 64.9 & 69.5 & 55.8 & 69.3 & 60.0 & 50.4 & 58.0 & 66.3 \\
XLM \cite{conneau2020unsupervised} &  85.0 &  78.7 &  78.9 &  77.8 &  76.6 &  77.4 &  75.3 &  72.5 &  73.1 &  76.1 &  73.2 &  76.5 &  69.6 &  68.4 &  67.3 &  75.1 \\
XLM (w/o TLM) \cite{conneau2020unsupervised}  & 83.2 & 76.7 & 77.7 & 74.0 & 72.7 & 74.1 & 72.7 & 68.7 & 68.6 & 72.9 & 68.9 & 72.5 & 65.6 & 58.2 & 62.4 & 70.7 \\
Unicoder \cite{huang2019unicoder}& 82.9 & 74.7 & 75.0 & 71.6 &71.6 & 73.2 & 70.6 & 68.7 & 68.5 & 71.2 & 67.0 & 69.7 & 66.0 & 64.1 & 62.5 & 70.5 \\
AMBER \cite{hu2020explicit} & 84.7 & 76.6 & 76.9 & 74.2 & 72.5 & 74.3 & 73.3 & 73.2 & 70.2 & 73.4 & 65.7 & 71.6 & 66.2 & 59.9 & 61.0 & 71.6\\
XLM-R$_{base}$ \cite{conneau2020unsupervised} & 84.6 & 78.2 & 79.2 & 77.0 & 75.9 & 77.5 & 75.5 & 72.9 & 72.1 & 74.8 & 71.6 & 73.7 & 69.8 & 64.7 & 65.1 & 74.2 \\
\hline 
MEP ({\it avg}) & 84.9 & 78.5 & 78.8 & 77.0 & 76.2 & 78.1 & 76.1 & 73.4 & 72.0 & 75.2 & 72.4 & 74.7 & 69.8 & 65.7 & 66.0 & 74.6 \\
OE ({\it avg}) & 84.4 & 78.1 & 78.8 & 77.1 & 75.9 & 78.0 & 75.9 & 73.1 & 72.5 & 75.3 & 73.0 & 74.5 & 70.1 & 65.4 & 67.3 & 74.6 \\
\hline 
\textbf{XLM-K} ({\it avg}) & 84.5 & 78.2 &  78.8 &  77.1 &  76.2 &  78.2 &  76.1 &  73.3 &  72.5 &  75.7 &  72.8 &  74.9 &  70.3 & 65.7 & 67.4 & 74.8 \\
\textbf{XLM-K} ({\it max}) & 84.9 &  79.1 &  79.2 &  77.9 & 77.2 &  78.8 &  77.4 &  73.7 &  73.3 & 76.8 &  73.1 &  75.6 &  72.0 &  65.8 &  68.0 &  \textbf{75.5} \\
\hline  
\hline
{\it with Bilingual Data} \\
\hline 
InfoXLM \cite{chi2020infoxlm}& 86.4  & 80.6 & 80.8 & 78.9 & 77.8 & 78.9 & 77.6 & 75.6 & 74.0 & 77.0 & 73.7 & 76.7 & 72.0 & 66.4 & 67.1 & 76.2 \\
ERNIE-M \cite{ouyang2020ernie}& 85.5 & 80.1 & 81.2 & 79.2 & 79.1 & 80.4 & 78.1 & 76.8 & 76.3 & 78.3 & 75.8 & 77.4 & 72.9 & 69.5 & 68.8 & \textbf{77.3} \\
\hline 
\end{tabular}}
\caption{The results of XNLI test accuracy on 15 languages. The models in the second block are our ablation models. The meaning of {\it avg} and {\it max} are the same as Table  \ref{table:mlqa}. Statistical significance test (Standard Deviation $\sigma$, {\it value}$\pm\sigma$) for XLM-K({\it avg}) on Avg: 74.8$\pm$0.3.}
\label{table:xnli}  
\end{table*}

\subsection{Ablation Study}

The ablation analysis mentioned above demonstrates the superiority of the combination scheme of the proposed two pre-training tasks. In this section, we investigate the  effectiveness of our key components. 

\begin{table}[t]
\centering
\resizebox{0.45\textwidth}{!}{
\begin{tabular}{l|ccc}
\hline
Model & MLQA & NER  & XNLI \\ \hline
XLM-R$_{base}$ & 65.1 & 79.0 & 74.2 \\ \hline
XLM-K w/o knowledge tasks ({\it avg}) & 65.6 & 79.0 & 74.5 \\
\hline
MEP w/o multilingual description ({\it avg})  & 65.9 & 79.3 & 74.5 \\
MEP ({\it avg})    & 66.4 & 79.6 & 74.6 \\ \hline
OE w/o contrastive loss ({\it avg}) & 65.7 & 79.6 & 74.5 \\
OE ({\it avg})    & 66.7 & \textbf{79.9} & 74.6 \\ \hline
\textbf{XLM-K} ({\it avg})   & \textbf{67.1} & 79.8 & \textbf{74.8} \\ \hline
\end{tabular}}
\caption{The ablation results on MLQA (F1 scores), NER (F1 scores) and XNLI (test accuracy) upon Avg. We run our model and ablation models four times with different seeds, where {\it avg} means the average results.}
\label{label:ablation}
\end{table}

\noindent \textbf{The Effectiveness of Knowledge Tasks} Our baseline model XLM-R$_{base}$ \cite{conneau2020unsupervised} was trained on Common Crawl dataset \cite{wenzek2019ccnet}, which covers our training data Wikipedia. As shown in Table \ref{table:mlqa}, \ref{table:ner}, \ref{table:xnli} and \ref{table:probe}, our model XLM-K outperforms XLM-R$_{base}$ consistently. Moreover, we replace $\mathcal{L}_e$ and $\mathcal{L}_o$ in Eq. \ref{eq:loss} with the MLM loss on multilingual Wikipedia entity descriptions. The results are shown in the second block of Table \ref{label:ablation}. Without the knowledge tasks, the performance of XLM-K w/o knowledge tasks drops by 1.5, 0.8 and 0.3 on MLQA, NER and XNLI respectively. It proves that the improvements are from the designed knowledge tasks, rather than the domain adaptation to Wikipedia. We will design more knowledge-related tasks in the future.


\noindent \textbf{The Effectiveness of Multilingual Entity Description} 
As mentioned in Sec. \ref{know-cons}, the entity knowledge, i.e. the entity-related Wikipedia page, is converted to different language resources compared to the given entity. The same operation is conducted on the {\it subject} and {\it object} in triplets, leading to the multilingual resources between the {\it subject} and the {\it object}. To study how this operation affects model performance, we report the results on the 
third block of Table \ref{label:ablation}. 
Without multilingual entity description operation, the performance of MEP w/o multilingual description drops by 0.5, 0.3 and 0.1 on MLQA, NER and XNLI respectively. It illustrates that the effectiveness of multilingual entity description. On the other hand, compared to baseline XLM-R$_{base}$, the model MEP w/o multilingual description still achieves 0.8, 0.3 and 0.3 improvements on MLQA, NER and XNLI, respectively, which reflects that applying entity description expansion without cross-lingual information in pre-training is still consistently effective for all downstream tasks. 


\noindent \textbf{The Effectiveness of Optimization Strategy} 
A natural idea to introducing structural knowledge into the pre-trained language model is to classify the {\it relation} by the {\it subject} and {\it object} from the triplets. Motivated by this opinion, we display the results on the forth block in Table \ref{label:ablation}.
The model of OE w/o contrastive loss classifies the {\it relation} by the concatenation of {\it subject} and {\it object} with Cross Entropy loss. Without contrastive loss, the performance drops by 1.0, 0.3 and 0.1 on MLQA, NER and XNLI respectively. This indicates the advantages of utilizing the contrastive learning towards a better cross-lingual model. We conjecture contrastive loss introduces a more challenging task than classification task. On the other hand, OE w/o contrastive loss improves the performance of baseline model XLM-R$_{base}$ from 65.1 to 65.7, 79.0 to 79.6 and 74.2 to 74.5 in MLQA, NER and XNLI, respectively. This observation certifies the importance of the structure knowledge in cross-lingual pre-training, though via an ordinary optimization strategy.

\begin{table*}[t]
\centering
\resizebox{0.9\textwidth}{!}{
\begin{tabular}{lcccccccc}
\hline
\multirow{2}{*}{Corpus}    & \multirow{2}{*}{Relation} & \multicolumn{2}{c}{Statistics} & XLM-R$_{base}$ & XLM-K w/o K & MEP  & OE & XLM-K \\ \cline{3-9} 
                           &                           & \# Facts        & \# Rel       & P@1        & P@1                               & P@1                                    & P@1      & P@1   \\ \hline
\multirow{4}{*}{Google-RE} & birth-place               & 2937            & 1            & 9.3    & 9.8    & 10.1                              &                         15.0             & \textbf{15.6}       \\
                           & birth-date                & 1825            & 1            & 0.6    & 0.7    & 0.7                               &          0.9                              & \textbf{1.0}        \\
                           & death-place               & 765             & 1            & 8.0     & 9.1    & 13.4                              &       13.8                               & \textbf{17.0}       \\ \cline{2-9} 
                           & Total                     & 5527            & 3            & 7.4     & 7.8    & 8.0                               &         9.9                               & \textbf{11.2}       \\ \hline
\multirow{4}{*}{T-REx}     & 1-1                       & 937             & 2            & 48.4    & 49.9    & 52.5                              &                 50.5                       & \textbf{62.0}       \\
                           & N-1                       & 20006           & 23           & 22.0   &  25.1    & 27.3                              &           21.9                             & \textbf{29.4}       \\
                           & N-M                       & 13096           & 16           & 17.9    & 21.5    & 25.6                              &        22.1                                & \textbf{26.1}       \\ \cline{2-9} 
                           & Total                     & 34039           & 41           & 21.7     & 22.8   & 27.9                              &         23.4                               & \textbf{29.7}       \\ \hline
ConceptNet                 & Total                     & 11458           & 16           & \textbf{18.8}   & 14.2     & 12.0                              &     17.6              & 15.7       \\ \hline
SQuAD                      & Total                     & 305             & -            & 5.5     & 6.4    & 10.1                              &                            9.7            & \textbf{11.5}       \\ \hline
\end{tabular}
}
\caption{The results of LAMA probing mean precision at one (P@1) for the baseline XLM-R$_{base}$, XLM-K w/o K (XLM-K w/o knowledge tasks), MEP, OE and XLM-K. We also reported the statistics of the facts number and relation types involved by the referred corpus.}
\label{table:probe}  
\end{table*}

\subsection{Probing Analysis}
We conduct a knowledge-aware probing task based on LAMA \cite{petroni2019language}. Note that the Probing is an \textbf{analysis experiment} to evaluate how well the pre-trained language model can store the desired (Wiki) knowledge, and to explain the reason for the improvements on downstream tasks by the proposed tasks. It means that the probing is not the SOTA comparison experiment.
We leave the analysis on recent multilingual LAMA \cite{jiang-etal-2020-x,kassner-etal-2021-multilingual} as our future work.



In LAMA, factual knowledge, such as $<$\textit{Jack, born-in, Canada}$>$, is firstly converted into cloze test question, such as ``\textit{Jack was born in \underline{~~~~~} }''. Then, a pre-trained language model is asked to predict the answer by filling in the blank of the question. There are 4 sub-tasks in the LAMA dataset. The first is \textbf{Google-RE}, which contains questions generated based on around 60k facts extracted from Wikidata and covers 3 relations. The second is \textbf{T-REx}, which contains questions generated based on a subset of Wikidata triples as well, but covers more relations (i.e. 41 relations in total). The third is \textbf{ConceptNet}, which contains questions generated based on a commonsense knowledge base \cite{speer2017conceptnet}. The last is a popular open-domain question answering dataset \textbf{SQuAD}. The number of facts and relation types covered by the each sub-task are shown in the Table \ref{table:probe} column {\it Statistics}.



\noindent \textbf{Evaluation on LAMA Probing Task}
The LAMA probing task is conducted on the baseline model XLM-R$_{base}$, our two ablation models MEP (Masked Entity Prediction + MLM) and OE (Object Entailment + MLM),  XLM-K w/o knowledge tasks, and our full model XLM-K. The results are shown in Table \ref{table:probe}. 

\setlength{\parskip}{5pt}
$\bullet$ {\bf Comparison Results}

The XLM-K w/o knowledge tasks improves the performance slightly (in Google-RE, T-REx and SQuAD). It proves the improvements are from the designed tasks, rather than the domain adaptation to Wikipedia. We will detail the observations of the results on each corpus.

{\it Google-RE} XLM-K outperforms all the other models by a substantial margin, especially the baseline model XLM-R$_{base}$. It is worth noting that the two ablation models, namely MEP and OE in Table \ref{table:probe}, realizes 0.6 and 2.5 gain respectively, which proves each knowledge-aware pre-training task can independently help pre-trained models to embed factual knowledge in a better way.

{\it T-REx} This task contains more facts and relations compared to Google-RE. 
XLM-K boosts the Total metrics from 21.7 to 29.7. The model MEP and model OE improves the scores by 6.2 and 1.7, respectively. These results further demonstrate the effectiveness of XLM-K on knowledge-aware tasks.

{\it ConceptNet} The ConceptNet corpus calls for the commonsense knowledge, which is a different knowledge source from Wikipedia. In this work, we mainly take Wikipedia knowledge into consideration, which can explain the worse performance on ConceptNet. Extending our model to capture more knowledge resources, such as commonsense knowledge, is our future work. Meanwhile, we notice that the performance of model OE decreases slightly compared to model MEP and XLM-K. The reason for this phenomenon may lie in that the ConceptNet is collected as the triplets-style and the relation prediction task has a great skill to handle the relation structure knowledge.

{\it SQuAD} To investigate the performance of our model on open-domain cloze-style question answering corpus, we further evaluate the results on SQuAD. Again, our model achieves a great success on SQuAD. In detail, XLM-K achieves 11.5, which has a 6.0 gain over XLM-R$_{base}$.

\setlength{\parskip}{5pt}
$\bullet$ {\bf  Case Study } 

To make the analysis more explicit, we conduct case study in the LAMA probing corpus. As shown in Table \ref{table:case}, we study four cases. Take the last two cases for example, to fill in the blank of ``\textit{Gnoccchi is a kind of \underline{~~~~~}.}'', XLM-R$_{base}$ fails to answer the question, while XLM-K successfully accomplishes the blank with ``\textit{food}''. In the last case ``\textit{Tasila Mwale (born \underline{~~~~~}).}'', XLM-R$_{base}$ has no idea towards this fact and only predicts the answer with ``\textit{in}''  to complete the phrase ``\textit{born in}''. XLM-K answers this question excellently via the prediction of ``\textit{1984}''. It confirms that the XLM-K is indeed equipped with more specific knowledge.






\begin{table}[t]
\centering
\resizebox{0.468\textwidth}{!}{
\begin{tabular}{c|c|c|c}
\hline
\multirow{2}{*}{Cloze Statement}                             & Subject Label & \multirow{2}{*}{XLM-R$_{base}$} & \multirow{2}{*}{XLM-K} \\ \cline{2-2}
                                                             & Object Label  &                            &                             \\ \hline
\multirow{2}{*}{Sometimes daydreaming causes an \underline{~~~~~}.} & daydreaming   & \multirow{2}{*}{alarm}     & \multirow{2}{*}{accident}   \\ \cline{2-2}
                                                             & accident      &                            &                             \\ \hline
\multirow{2}{*}{Phones   may be made of  \underline{~~~~~}.}        & phones        & \multirow{2}{*}{metal}     & \multirow{2}{*}{plastic}    \\ \cline{2-2}
                                                             & plastic       &                            &                             \\ \hline
\multirow{2}{*}{Gnocchi   is a kind of  \underline{~~~~~}.}         & gnocchi       & \multirow{2}{*}{beer}      & \multirow{2}{*}{food}       \\ \cline{2-2}
                                                             & food          &                            &                             \\ \hline
\multirow{2}{*}{Tasila Mwale (born    \underline{~~~~~}).}          & Tasila Mwale  & \multirow{2}{*}{in}        & \multirow{2}{*}{1984}       \\ \cline{2-2}
                                                             & 1984          &                            &                             \\ \hline
\end{tabular}}
\caption{Case study of LAMA probing, where the object label is the ground truth of the given statement. We compare the prediciton from our baseline XLM-R$_{base}$ and our full model XLM-K. }
\label{table:case}  
\end{table}

\section{Conclusion}

In this work, we present a new cross-lingual language model XLM-K to associate pre-training language model with more specific knowledge across multiple languages. Specifically, the knowledge is obtained via two knowledge-related tasks: maksed entity prediction and object entailment. Experimental results on three benchmark datasets clearly demonstrate the superiority of XLM-K. Our systematic analysis of the XLM-K advocates that XLM-K has great advantages in knowledge intensive tasks. Incorporating more diverse multilingual knowledge and jointing more advanced pre-training schemes will be addressed in future work.


\bibliography{aaai22} 

\clearpage
\begin{appendices}

\section{Details of Knowledge Construction}

\begin{figure*}[t]
\centering
\includegraphics[width=16cm]{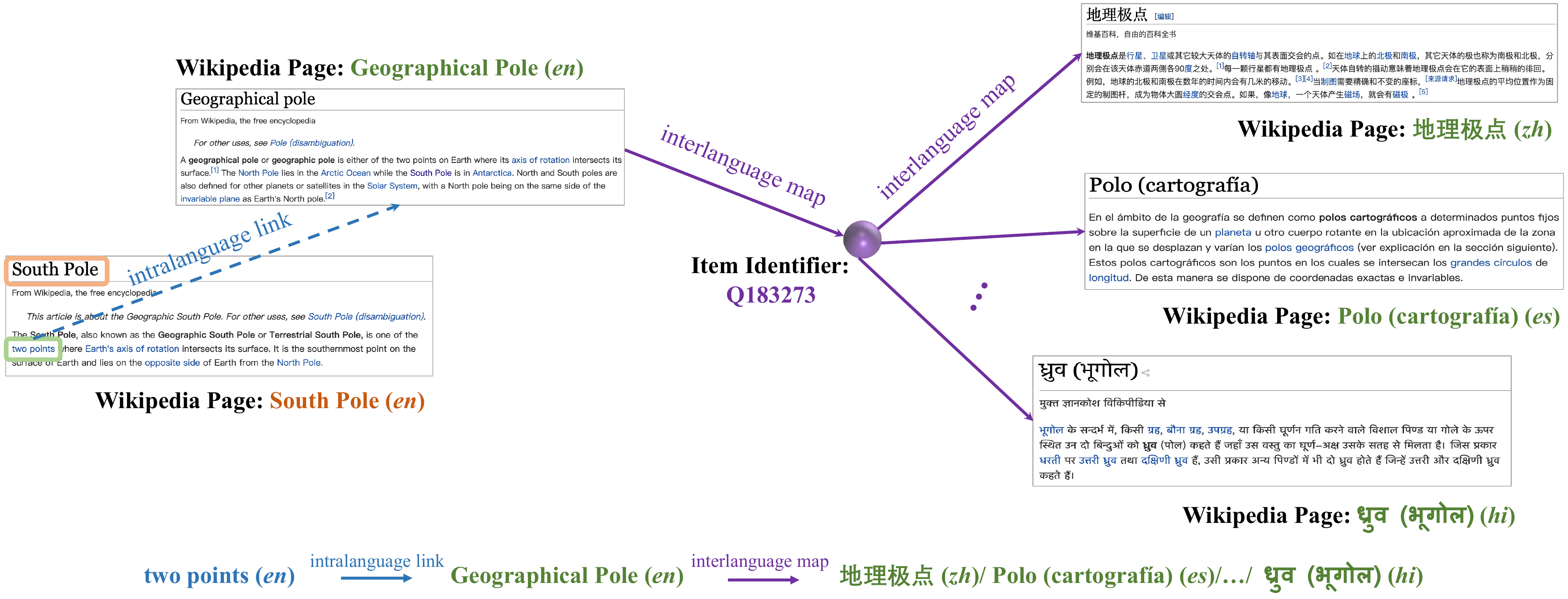}
\caption{An example of cross-lingual entity knowledge. Each mentioned entity is labeled with blue color and is linked to its related page by intralanguage link. Then each page is viewed as the entity in Wikdata, in which the same entity in different languages is linked with item identifier.}
\label{pic:entity}
\end{figure*}

\noindent \textbf{Entity Mention} As shown in Figure \ref{pic:entity}, the Wikipedia page {\it South Pole}\footnote{\url{https://en.wikipedia.org/wiki/South_Pole}} is labeled with mentioned entity {\it two points} and {\it Earth’s axis of rotation} etc. Take entity {\it two points} for example, it is linked to its referred Wikipedia page {\it Geographical Pole}\footnote{\url{https://en.wikipedia.org/wiki/Geographical_pole}} by intralanguage link. The contextual description of Geographical Pole is the monolingual knowledge of mentioned entity {\it two points}.

\begin{figure}[t]
\centering
\includegraphics[width=7cm]{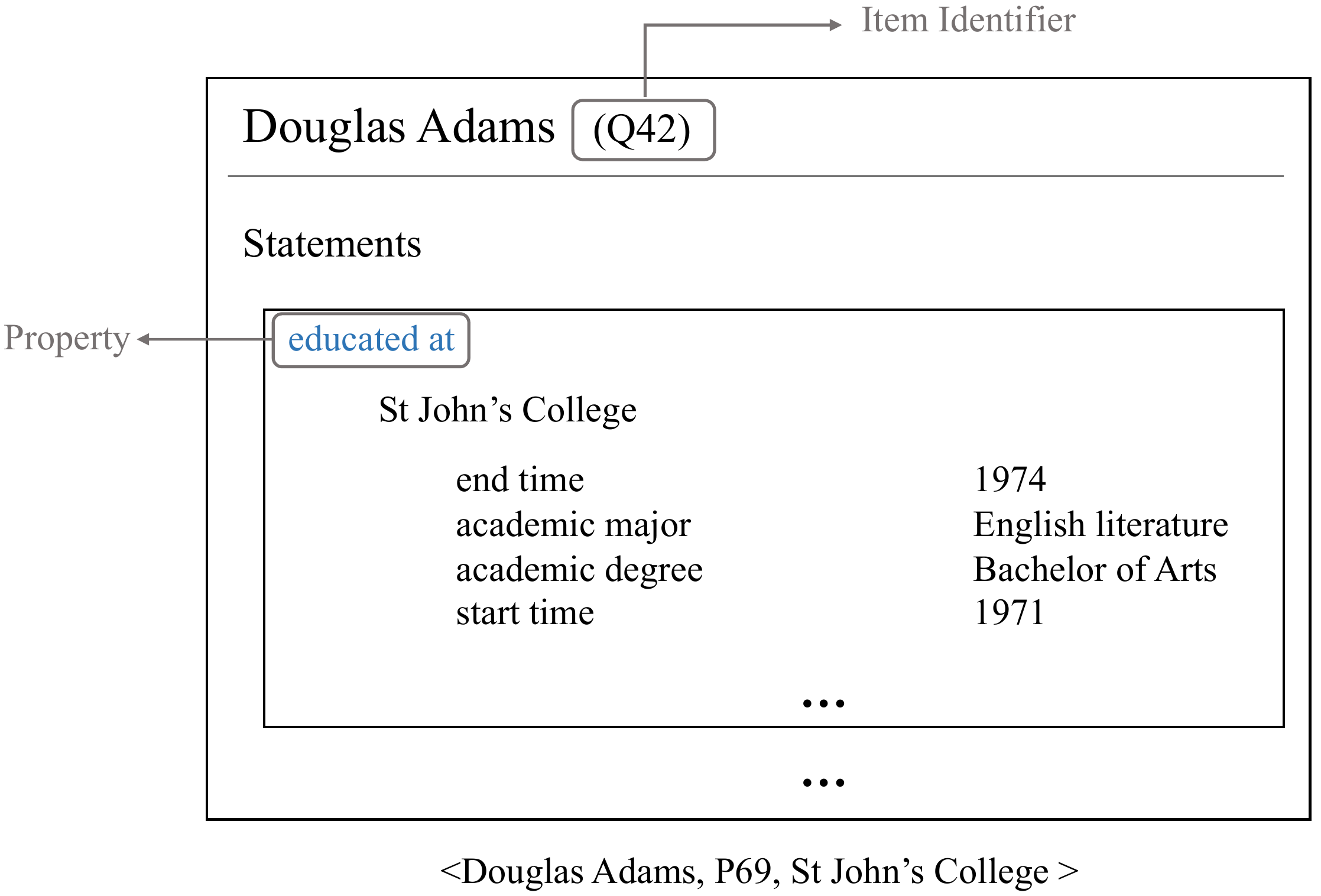}
\caption{An example of relation knowledge construction from Wikidata. In Wikidata, Property is viewed as relation, where {\it educated at} is the Property label and P69 is the corresponding relation ID.}
\label{pic:rel}
\end{figure}

\noindent \textbf{Knowledge Graph} We utilize Wikidata to construct relation knowledge. As shown in Figure \ref{pic:rel}\footnote{\url{https://www.wikidata.org/wiki/Wikidata:Glossary}}, the {\it Douglas Adams} and {\it St John’s College} are two Wikipedia pages, which are viewed as entities in Wikidata. The property\footnote{\url{https://www.wikidata.org/wiki/Help:Properties}} is the relation between the two entities, which is presented as relation ID in the Wikidata. The mapping of relation ID to property label can be found in the referred page\footnote{\url{https://www.wikidata.org/wiki/Wikidata:Database_reports/List_of_properties/all}}. Then, we can get the triplets $<$Douglas Adams, P69, St John’s College$>$.

\noindent \textbf{Multilingual Entity Description} Wikidata\footnote{\url{https://www.wikidata.org}} contains the cross-lingual information via title mapping and each Wikipedia page is an entity in WikiData. As shown in Figure \ref{pic:entity}, each entity has a typical item identifier ID, e.g. Q183273 is the ID for entity {\it Geographical Pole}. By the typical item identifier ID, the same entity can be associated across the different languages. Finally, the entity {\it two points} can be linked with multilingual long entity description via intralanguage link and interlanguage map. The entity description in the triplets can also be converted to multilingual style by this manner.


\section{Knowledge Task Analysis}

To figure out whether our model can meet the demands of knowledge task, we further propose two knowledge-based tasks in cross-lingual domain: {\it cross-lingual entity prediction} and {\it cross-lingual relation recognition}. Then the comparison is conducted on these two tasks.

\noindent \textbf{Definition of Knowledge Task}

{\it Cross-lingual Entity Prediction (XEP)} aims to predict the entity description for the masked token given a sentence. The input is the entity-masked sentence, like ``{\ttfamily [Mask]} have yellow skin''. The task is to select the best description from the candidate description list, where the list is language agnostic and the list capacity is 32k. When the correct answer is ranked on the first position, we record the scores.

{\it Cross-lingual Object Recognition (XOR)} targets on the recognition of object from the description of subject and relation. The input is the sentence of subject and relation. The task is to rank the candidate object description list based on the add operation of subject and relation. When the correct answer is ranked on the first position, we record the scores. Again, the options in the candidate list are language agnostic and the list capacity is 32k.

\begin{figure}[t]
\centering
\includegraphics[width=7cm]{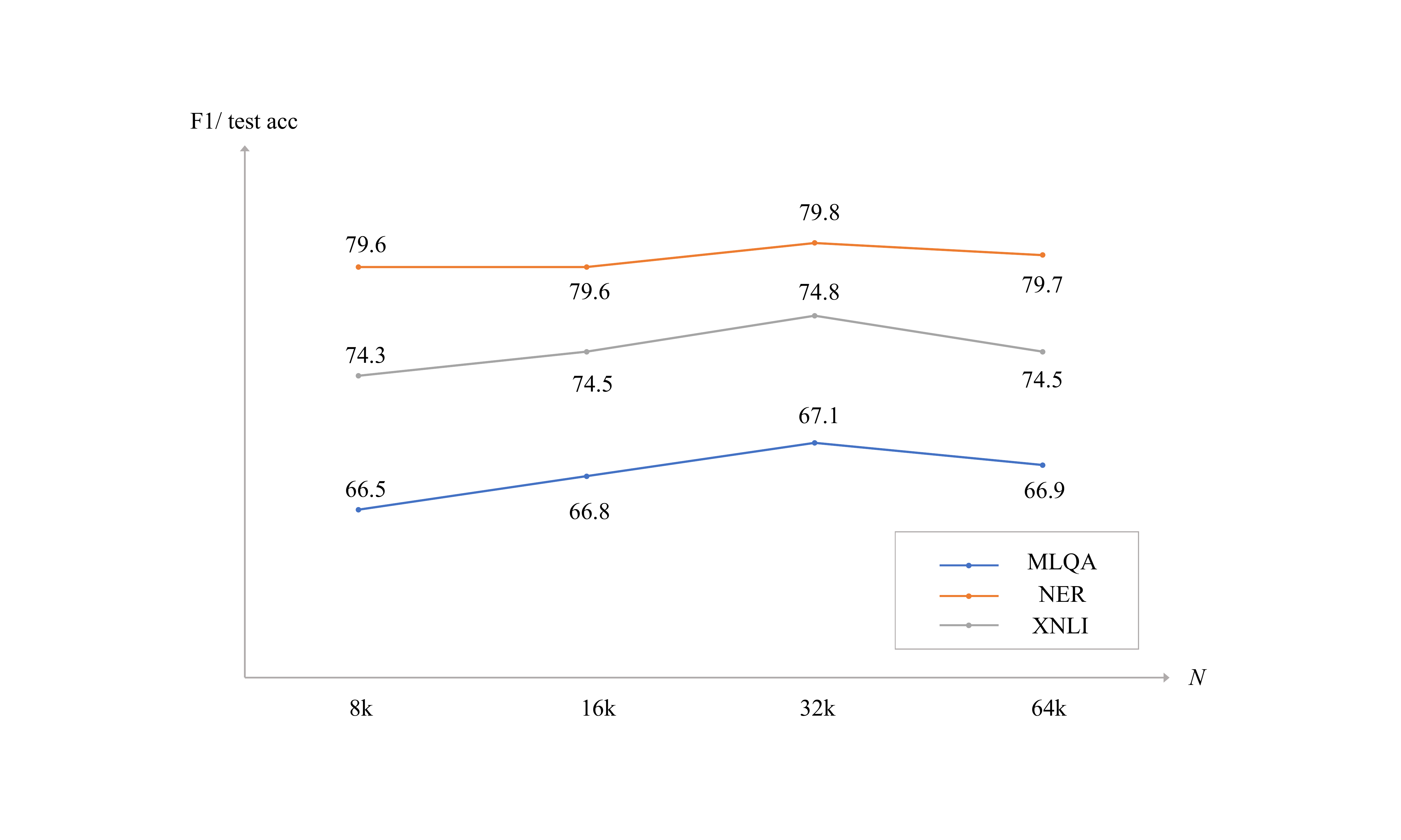}
\caption{The results of different size $N$ of the Candidate List $\mathcal{Z}$. We select $N=8$k$,16$k$,32$k and $64$k and report the F1 scores on MLQA, F1 scores on NER and test accuracy on XNLI.}
\label{pic:mbsize}
\end{figure}

\noindent \textbf{Evaluation on Knowledge Task}

We conduct experiments on the examine task XEP and XOR to further evaluate our model. The results are shown in Table \ref{table:know}. The main observation is concluded as follows:

(1) {\it The Capability of Entity and Relation Recognition.} The XLM-K boosts the performance of XEP and XOR, achieving 2.1 and 1.5 individually. The 2.0 and 1.49 gain, compared to baseline XLM-R, proves that our model has great ability to capture the entity and relation knowledge.

(2) {\it The Learning Mode of the Combination of the Two Pre-Trainging Tasks.} We notice that the XLM-K performs best on XEP and turns a little lower on XOR than the ablation models. It reveals the learning mode of XLM-K after the combination the two designed tasks, namely the Object Entailment Task help the model learn the entity knowledge better and the relation information stored in the model will be weaken slightly. In addition, the 1.5 performance on XOR also achieves a great margin than the baseline on XOR. These observations certify that the entity knowledge and relation knowledge are learned via the proposed pre-training task in succeed.

\section{Details of the Candidate List}

\begin{table}[t] 
\centering
\resizebox{0.4\textwidth}{!}{
\begin{tabular}{lcc}
\hline                       
Model & XEP & XOR \\ 
\hline 
XLM-R & 0.1 & 0.01 \\
\hline
MEP (Masked Entity Prediction + MLM) & 1.1 & -\\
OE (Object Entailment + MLM) & - & \textbf{2.0} \\
\hline 
\textbf{XLM-K}  & \textbf{2.1} & 1.5 \\
\hline  
\end{tabular}}  
\caption{The results of XEP and XOR accuracy scores. The first block is our baseline and the second block is our ablation models.}
\label{table:know}  
\end{table}

In this section, we will introduce the influence of Candidate List size $N$ and detailed implementation of the Candidate List $\mathcal{Z}$. 

\noindent \textbf{The Influence of Candidate List Size}

In ideal situation, the Candidate List should contain all the entities from the data. However, due to the limitation of device memory, the size $N$ of Candidate List is not unlimited. To figure out the influence of $N$, we conduct experiments on MLQA, NER and XLNI, when $N=8$k$,16$k$,32$k and $64$k. As shown in Figure \ref{pic:mbsize}, when $N=32$k, the model achieves the best results on MLQA, NER and XNLI. Thus, the $N$ is set as 32k in our main experiments.

\noindent \textbf{The Implementation of Candidate List} 

The Candidate List is implemented as a queue. There exist two key operations about the Candidate List $\mathcal{Z}$: {\it initialization} and {\it updating}.

\textbf{Initialization} At the beginning of the training stage, the Candidate List $\mathcal{Z}$ is randomly initialized, denotes  $\mathcal{Z} =\{Z_1^r, Z_2^r, ..., Z_N^r\}$, where $Z_i^r$ ($i=1,2,..,N$) is the $i$-th initialized sample in $\mathcal{Z}$.

\textbf{Updating} At each training step  $\tau$, we can get the positive sample ${Z_{+\tau}}$ of the input data. Then the $\mathcal{Z}$ is updated by ${Z_{+\tau}}$. Hypothesis each training step only has one training data, the Updating process can be described as:

(1) Training step 1: The positive sample is ${Z_{+1}}$ and $\mathcal{Z}$ is updated by ${Z_{+1}}$, then $\mathcal{Z}=\{ Z_2^r, Z_3^r, ..., Z_{N}^r , Z_{+1}\}$;

(2) Training step 2: The positive sample is ${Z_{+2}}$ and $\mathcal{Z}$ is updated by ${Z_{+2}}$, then $\mathcal{Z}=\{Z_3^r, Z_4^r, ..., Z_{+1}, Z_{+2}\}$;

(3) Training step 3: The positive sample is ${Z_{+3}}$ and $\mathcal{Z}$ is updated by ${Z_{+3}}$, then $\mathcal{Z}=\{Z_4^r, Z_5^r, ..., Z_{+2}, Z_{+3}\}$;

......

\noindent At each training step $\tau$, there is only one positive sample $Z_{+\tau}$ in $\mathcal{Z}$  and others are negative samples.

\section{Case Study of MLQA}

\begin{figure*}[t]
\centering
\includegraphics[width=16cm]{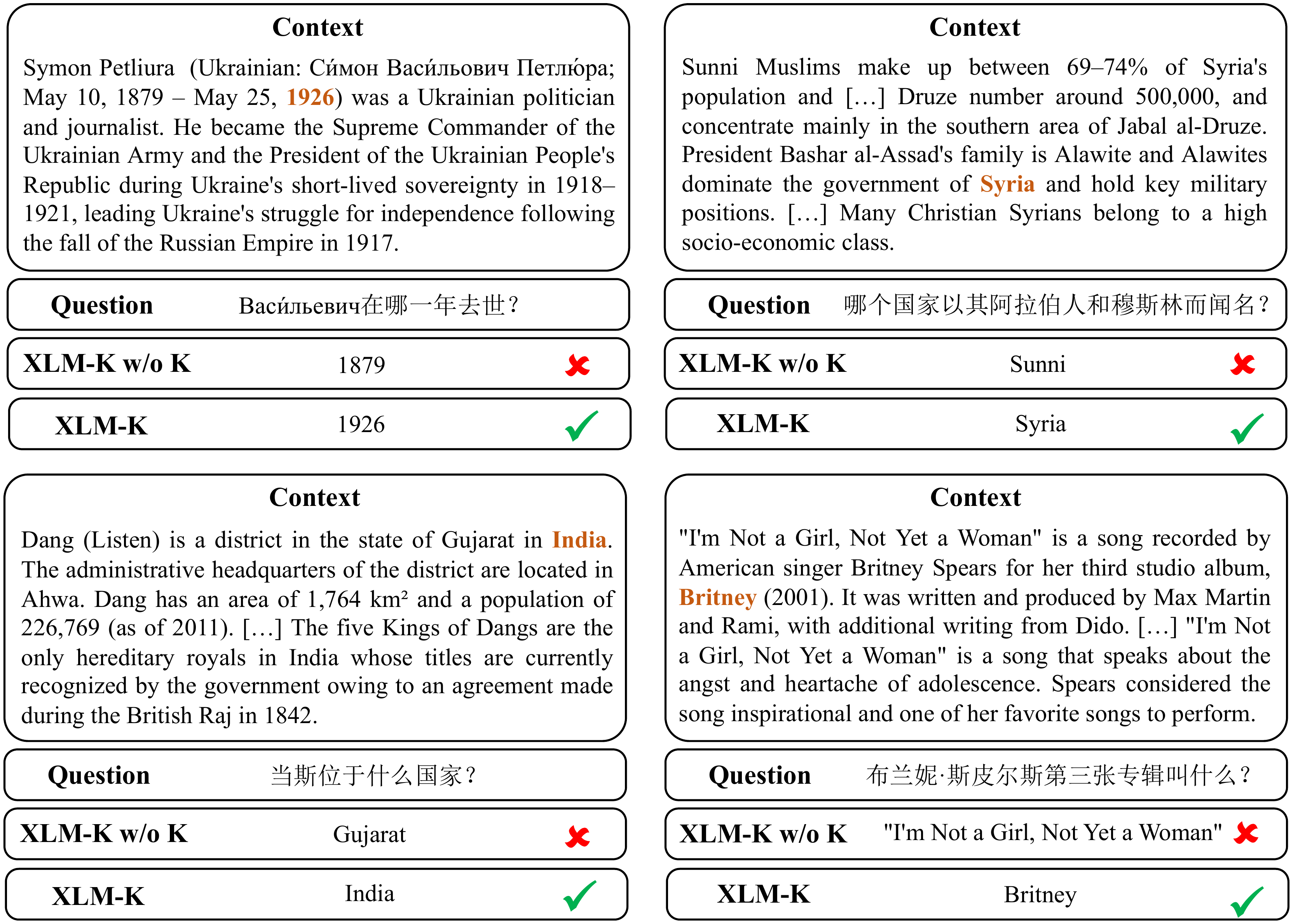}
\caption{Case study of the results of MLQA. We report the answer predicted by the XLM-K and XLM-K w/o K (XLM-K w/o knowledge tasks) respectively. The context and the answer are in English ({\it en}), while the question is in Chinese ({\it zh}). The ground truth is highlighted with orange color in the context. Context shortened for clarity with ``[...]''.}
\label{pic:mlqa-case}
\end{figure*}

To figure out what type of improvement is provided by our model XLM-K, we conduct Case Study on MLQA task. As shown in Figure \ref{pic:mlqa-case}, we display four cases for analysis.

For the first case, the question is ``What year did Symon Petliura die?''. Our model XLM-K answers this question correctly, while XLM-K w/o K answers with his year of birth ``1879''.

For the second case, the question is ``Which country is famous for Arabs and Muslims?''. Our model XLM-K answers with ``Syria'' successfully, while XLM-K w/o K answers with ``Sunni''.

For the third case, the question is ``In which country is Dang located?''. Our model XLM-K answers this question in succeed with ``India''. However, XLM-K w/o K answers with ``Gujarat'', which is a state on the western coast of India.

For the last case, the question is ``What is the name of Britney Spears' third album?''. ``Britney'' is the correct answer. ``I'm Not a Girl, Not Yet a Woman'' is the name of the song from the Britney rather than the name of album.

From the above analysis, we can find that the knowledge from the Wikipedia helps the model do better when dealing with the knowledge-related problems. The entity (like {\it Symon Petliura}) and the relationship between entities (like $<${\it Symon Petliura, date of death, May 25 1926}$>$) are learnt by the model via the proposed knowledge tasks.

\end{appendices}

\end{document}